\documentclass[twocolumn]{galbot}
\usepackage{wrapfig}
\usepackage{tabularx}
\usepackage{textcomp}
\usepackage{stfloats}
\usepackage{url}
\usepackage{verbatim}
\usepackage{graphicx}
\usepackage{titlesec}
\usepackage{tocloft}
\usepackage{adjustbox}
\usepackage{multirow}
\usepackage{tikz}
\usepackage{comment}
\usepackage{amsmath,amssymb} % define this before the line numbering.
\usepackage{colortbl}  %彩色表格需要加载的宏包
\usepackage{color}
\usepackage{booktabs} 
\usepackage{hyperref}
\usepackage{graphicx}    % 用于插入图片
\usepackage{subcaption} 
\usepackage{booktabs}
\usepackage{amsmath} % 数学公式支持
\usepackage{multirow} % 表格多行合并
\usepackage{booktabs} % 表格更好的线条
\usepackage{arydshln} % 虚线表格分割线
\usepackage{subcaption} % 支持 subtable 和 subfigure
\RequirePackage{xspace}
\makeatletter
\DeclareRobustCommand\onedot{\futurelet\@let@token\@onedot}
\def\@onedot{\ifx\@let@token.\else.\null\fi\xspace}

\makeatother

\usepackage{makecell}

\definecolor{adptorange}{RGB}{248, 205, 172}
\definecolor{cmpblue}{RGB}{189, 215, 238}
\definecolor{cmpblue}{RGB}{189, 215, 238}

\definecolor{our_red}{RGB}{232,157,160}
\definecolor{our_blue}{RGB}{136,206,230}
\definecolor{our_orange}{RGB}{246,200,168}
\definecolor{our_green}{RGB}{178,211,164}

\definecolor{attn_code0}{RGB}{247,215,200}
\definecolor{attn_code1}{RGB}{238,169,139}
\definecolor{mlp_code0}{RGB}{204,201,221}
\definecolor{mlp_code1}{RGB}{102,95,153}

\definecolor{linecolor1}{RGB}{246, 248, 239}
\definecolor{linecolor2}{RGB}{230, 234, 217}
\definecolor{linecolor3}{RGB}{211, 222, 190}

\definecolor{token_blue}{RGB}{84, 120, 140}
\definecolor{myMagenta}{rgb}{0.9,0,0.4}

\usepackage{pifont}       % \ding{xx}
\usepackage{bbding}       % \Checkmark,\XSolid,... (需要和pifont宏包共同使用)
\usepackage{fontawesome}
\usepackage{xspace}

\usepackage{float}

\newlength\savewidth

\newcolumntype{x}[1]{>{\centering\arraybackslash}p{#1pt}}
\newcolumntype{y}[1]{>{\raggedright\arraybackslash}p{#1pt}}
\newcolumntype{z}[1]{>{\raggedleft\arraybackslash}p{#1pt}}

\renewcommand{\paragraph}[1]{\vspace{1mm}\noindent\textbf{#1}}
\usepackage{colortbl}
\usepackage{xcolor}
\usepackage{wrapfig}
\usepackage{graphicx}
\usepackage{amssymb}
\usepackage{adjustbox}
\usepackage{colortbl}
\usepackage{xcolor}
\usepackage{multirow}
\usepackage{array}
\usepackage{makecell}
\usepackage{bbm}
\usepackage{collcell,xfp}
\usepackage{pgf}
\usepackage{tikz}
\usepackage[most]{tcolorbox}
\usepackage{csquotes}
\usepackage[noorphans,vskip=1em,leftmargin=1em]{quoting}
\usepackage{enumitem} 
\usepackage{tcolorbox} 
\usepackage{forest}
\usepackage{wrapfig}
\usepackage{caption}
\usepackage{subcaption}
\usepackage{longtable}
\usepackage[T1]{fontenc}

\renewcommand{\paragraph}[1]{\vspace{1.25mm}\noindent\textbf{#1}}

\usepackage{algorithm}
\usepackage{listings}

\definecolor{codeblue}{rgb}{0.25, 0.5, 0.5}
\definecolor{codekw}{rgb}{0.35, 0.35, 0.75}
\lstdefinestyle{Pytorch}{
    language = Python,
    backgroundcolor = \color{white},
    basicstyle = \fontsize{9pt}{8pt}\selectfont\ttfamily\bfseries,
    columns = fullflexible,
    aboveskip=1pt,
    belowskip=1pt,
    breaklines = true,
    captionpos = b,
    commentstyle = \color{codeblue},
    keywordstyle = \color{codekw},
}

\definecolor{green}{HTML}{009000}
\definecolor{red}{HTML}{ea4335}

\title{Learning Athletic Humanoid Tennis Skills from Imperfect Human Motion Data}

\author[1, 3, *]{Zhikai Zhang}
\author[1, 3, *]{Haofei Lu}
\author[1, 3, *]{Yunrui Lian}
\author[1, 3]{Ziqing Chen}
\author[1, 3]{Yun Liu}
\author[3]{Chenghuai Lin}
\author[1, 3]{Han Xue}
\author[3]{Zicheng Zeng}
\author[1, 3]{Zekun Qi}
\author[3]{Shaolin Zheng}
\author[3]{Qing Luan}
\author[5]{Jingbo Wang}
\author[1]{Junliang Xing}
\author[2, 3]{He Wang}
\author[1, 4, \dagger]{Li Yi}

\affiliation[1]{Tsinghua University}
\affiliation[2]{Peking University}
\affiliation[3]{Galbot Inc.}
\affiliation[4]{Shanghai Qi Zhi Institute}
\affiliation[5]{Shanghai AI Laboratory}

\contribution[*]{Equal Contribution}
\contribution[\dagger]{Corresponding author}

\code{\url{https://github.com/GalaxyGeneralRobotics/LATENT}}
\page{\url{https://zzk273.github.io/LATENT/}}

\abstract{
Human athletes demonstrate versatile and highly-dynamic tennis skills to successfully conduct competitive rallies with a high-speed tennis ball. However, reproducing such behaviors on humanoid robots is difficult, partially due to the lack of perfect humanoid action data or human kinematic motion data in tennis scenarios as reference. In this work, we propose \textbf{LATENT}, a system that \textbf{L}earns \textbf{A}thletic humanoid \textbf{TE}nnis skills from imperfect human motio\textbf{N} da\textbf{T}a. 
The imperfect human motion data consist only of motion fragments that capture the primitive skills used when playing tennis rather than precise and complete human-tennis motion sequences from real-world tennis matches, thereby significantly reducing the difficulty of data collection. Our key insight is that, despite being imperfect, such quasi-realistic data still provide priors about human primitive skills in tennis scenarios. With further correction and composition, we learn a humanoid policy that can consistently strike incoming balls under a wide range of conditions and return them to target locations, while preserving natural motion styles.
We also propose a series of designs for robust sim-to-real transfer and deploy our policy on the Unitree G1 humanoid robot. Our method achieves surprising results in the real world and can stably sustain multi-shot rallies with human players as shown in Figure~\ref{fig:teaser}. }

\makeatletter
\apptocmd{\mymaketitle}{%
  \vspace{10pt}
  \begin{center}
      \captionsetup{type=figure}
      \includegraphics[width=0.85\textwidth]{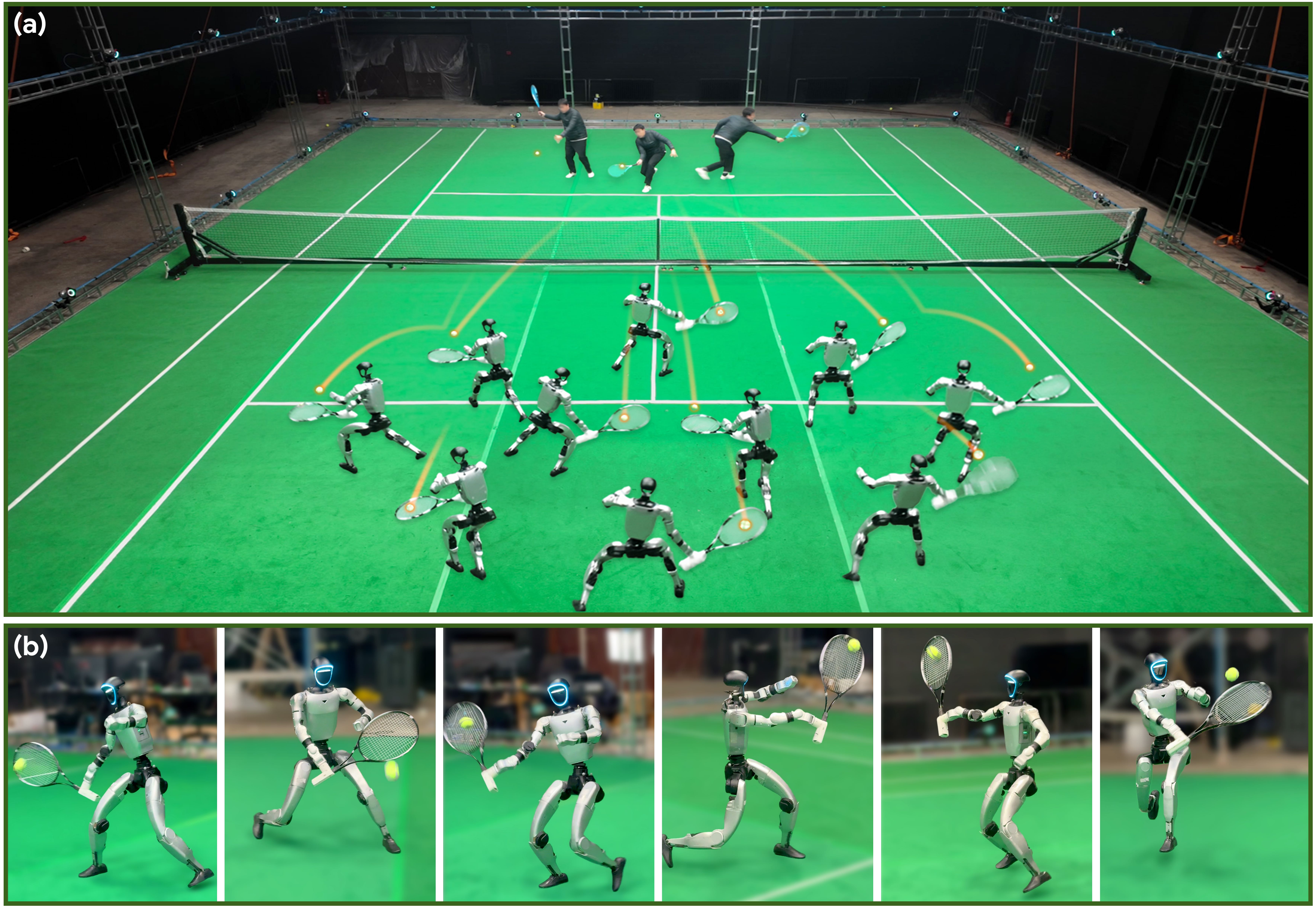}
      \vspace{10pt}
      \captionof{figure}{\textbf{(a)} The humanoid performs multi-shot rallies with a human player using different stroke types across various court regions. \textbf{(b)} The humanoid performs athletic tennis skills to strike an incoming ball traveling at high speed (peak velocities > \textit{15} m/s).}
      \label{fig:teaser}
  \end{center}
}{}{}
\makeatother

\begin{document}
\pagestyle{empty}
\maketitle

\section{Introduction}
Learning athletic sports skills remains one of the core challenges for humanoid robots. Such sports typically require highly dynamic motions, rapid reactions, and high precision. Taking tennis as an example, players are often required to sprint across the court at speeds exceeding \textit{6 m/s}, react to incoming balls traveling at \textit{15–30 m/s}, and strike the ball within an extremely short ball–racket contact duration of only \textit{a few milliseconds}.
These characteristics make it infeasible to collect humanoid action data via human-humanoid tele-operation for direct imitation learning, posing a significant challenge to reproducing athletic tennis skills on humanoids.

An alternative approach is to use kinematic human motion data as reference. However, tennis typically involves: \textbf{(i)}~large-scale human movements that span the half court, \textbf{(ii)}~sustained multi-rally interactions, and \textbf{(iii)}~subtle and highly precise wrist motions during ball striking. Such characteristics make it extremely difficult for motion capture systems to collect precise and complete human-tennis motion data from real-world matches. Vid2Player3D~\cite{zhang2023learning} alleviates this problem by leveraging broadcast videos. However, it needs a complex pipeline to extract human motion data from monocular videos, including player detection, camera estimation, 2D and 3D human pose estimation, followed by imitation policy with external residual forces to correct motion artifacts. Assembling these modules may require substantial expertise and engineering efforts.

In this work, we propose \textbf{LATENT}, a novel system to \textbf{L}earn \textbf{A}thletic humanoid \textbf{TE}nnis skills from imperfect human motio\textbf{N} da\textbf{T}a. The \textit{imperfect} human motion data consist only of motion fragments that capture the primitive skills (e.g., forehand stroke, backhand stroke, lateral shuffle, crossover step) used when playing tennis rather than realistic and complete human-tennis motion sequences from real-world tennis matches, thereby significantly reducing the difficulty of data collection.
Here, \textit{imperfect} manifests in two aspects: \ding{182}~\textbf{Imprecise}. Due to the motion capture difficulty and cross-embodiment gap, the wrist motions during racket swinging are often imprecise. \ding{183}~\textbf{Incomplete}. The motion data only provide priors about natural motor skills and do not contain any knowledge about how to utilize them for the tennis task. Our key insight is that, despite being imperfect, such quasi-realistic data adequately provide priors about human primitive skills in tennis scenarios. These knowledge, with further correction and composition, can be leveraged to learn tennis skills while maintaining natural human behavior.

Our method builds upon recent ideas in hierarchical humanoid control with latent action spaces: constructing a latent space to represent nature humanoid actions and then training a high-level policy to sample from it to accomplish the tennis return task. We propose two novel designs to mitigate the imperfections of the data. First, to \textit{bridge the gap between the high-precision nature of human-tennis interaction and the imprecise quasi-realistic human motion data}, we design the latent space to be compatible with potential action corrections predicted by the high-level policy. Second, since the motion data do not contain any knowledge about how these primitive skills should be utilized to return tennis, the high-level policy can easily exploit the latent space while learning to accomplish the task. To \textit{balance the task performance and adherence to the primitive skill priors}, we design a latent action barrier to constrain the RL exploration of the high-level policy according to the state-based action distribution prior.

To enable robust sim-to-real transfer, we also propose a series of careful designs for robust sim-to-real transfer, including dynamics randomization and observation noise applied to both the robot and the tennis ball.
We deploy our method on Unitree G1 hardware and can stably sustain multi-shot rallies with human players. 

Our contributions are fourfold:

\begin{itemize}
\item We propose \textbf{LATENT}, a novel system to learn athletic humanoid tennis skills from imperfect human motion data, paving the way for future research.
\item We propose two novel designs to better construct and utilize a latent action space from imperfect human motion data.
\item We successfully deploy our method on a real-world humanoid robot and can stably sustain multi-shot rallies with human players.
\item Extensive experiments are conducted in both simulation and the real world to validate the effectiveness of our major designs.

\end{itemize}

\section{Related Work}

\subsection{Latent Action Space for Humanoid Control} 
Due to the high degrees of freedom (DoFs) of humanoid robots, learning skills from scratch often leads to low RL sampling efficiency and unnatural motions. To address this problem, learning a latent action space that encodes reusable primitive skills from pre-recorded human motion data, and subsequently training a high-level policy to sample within this latent space, has been widely studied~\cite{zhang2023learning, zhang2025unleashing, yao2022controlvae, yao2024moconvq, luo2023universal, luo2024grasping, peng2022ase, tessler2023calm, dou2023c,ren2023insactor,won2022physics, li2025learning}. Some prior works~\cite{zhang2023learning,ren2023insactor} learn a kinematic motion latent space, followed by a pre-trained motion tracker to accomplish downstream tasks. However, the generated kinematic motions often exhibit physical infeasibility, posing significant challenges for the tracker. In contrast, some works~\cite{zhang2025unleashing, yao2022controlvae, yao2024moconvq, luo2023universal, peng2022ase, tessler2023calm, li2025learning} directly learn a latent space of joint actions. PULSE~\cite{luo2023universal} distills a VAE-style latent action space from a universal motion tracker. $\text{R}^{2}\text{S}^{2}$~\cite{zhang2025unleashing} adopts this idea on real humanoid robots and learns a latent space from a set of pre-trained primitive skills to solve loco-manipulation tasks.

Compared with existing works, we focus on learning athletic humanoid tennis skills from imperfect human motion data that contains only imprecise and incomplete primitive skills. We propose two key novel designs to correct and compose these imperfect human motion priors so that the high-level policy can learn to accomplish the task while preserving natural motion styles.

\subsection{Humanoid for Sports}
Humanoid for sports have long been a central topic in character animation. A significant amount of works ~\cite{zhang2023learning, luo2024smplolympics, wang2025skillmimic, yu2025skillmimic, xu2025learning, kim2025physicsfc, zhang2024freemotion, wang2024strategy, xu2025parc, liu2018learning} have been proposed to enable humanoids to play various sports, including tennis~\cite{zhang2023learning}, basketball~\cite{wang2025skillmimic, yu2025skillmimic, xu2025learning}, and football~\cite{kim2025physicsfc}. Among these, the work most closely related to ours is Vid2Player3D~\cite{zhang2023learning}, which builds a simulated character that can hit the incoming ball to target positions using a diverse array of strokes. However, this work relies on several physically infeasible assumptions, such as unlimited joint torques and residual force on the root~\cite{yuan2020residual}, and does not consider the sim-to-real gap of the overall system. In addition, our work focuses on how to leverage easily collected yet imperfect human motion data to learn tennis skills.

With recent advances in humanoid hardware, several works~\cite{haarnoja2024learning, su2025hitter, liu2025humanoid, hu2025towards, wang2025hierarchical, wang2025learning, yin2026robostriker, wang2026humanx, chen2026learning, kong2026learning, ren2025humanoid} successfully learn sports skills on a real-world humanoid robot, including table tennis~\cite{su2025hitter, hu2025towards}, badminton~\cite{liu2025humanoid, chen2026learning}, football~\cite{wang2025hierarchical, wang2025learning, kong2026learning}, and boxing~\cite{yin2026robostriker}. However, these works often lack highly dynamic motions, agile body coordination, or rapid reaction. Our work takes a significant step toward achieving athletic humanoid sports skills.

\section{Learning Tennis Skills with LATENT}

\begin{figure*}
    \centering
    \includegraphics[width=0.9\linewidth]{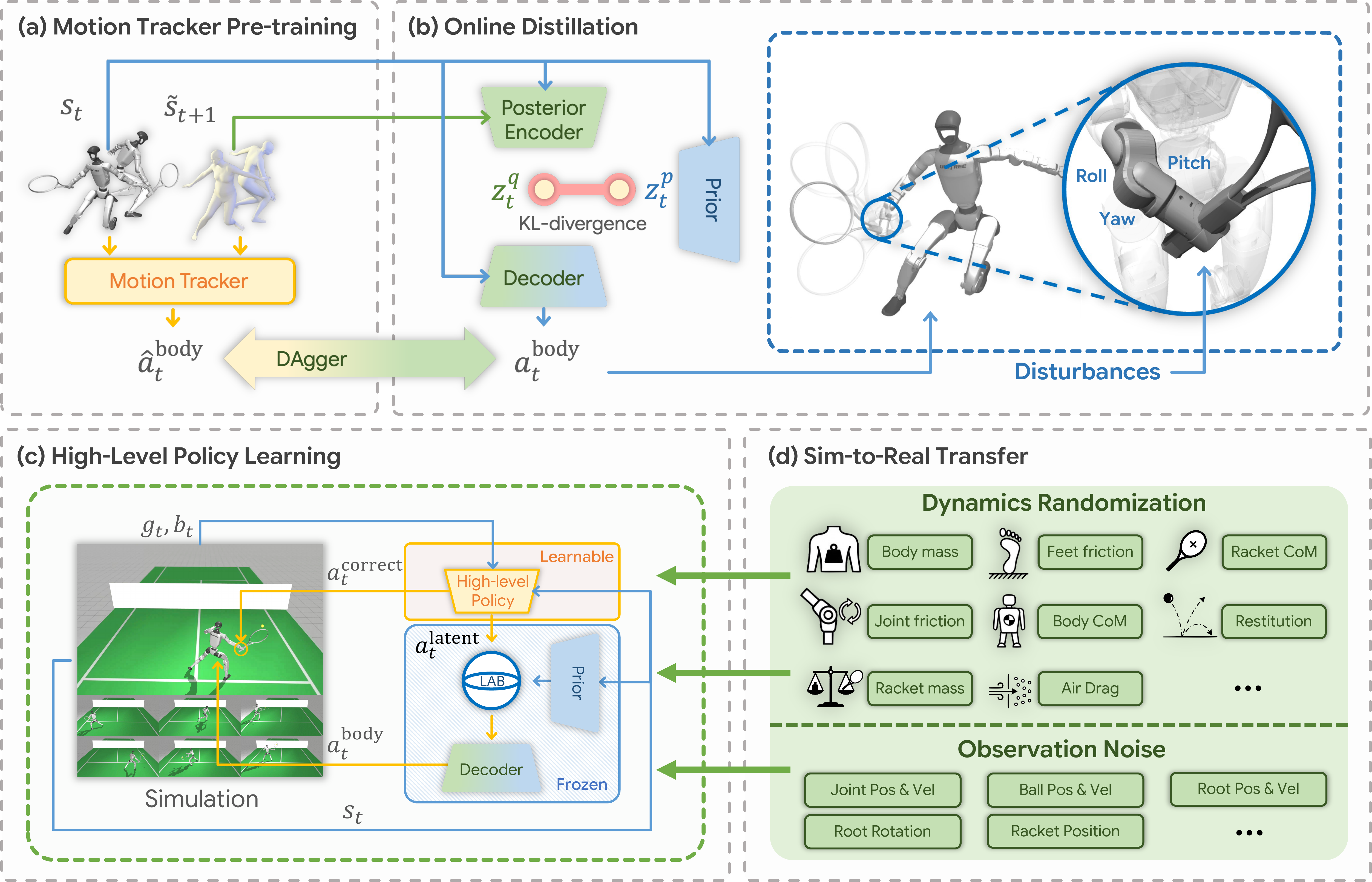}
    \vspace{0.2cm}
    \caption{\textbf{Overview of LATENT}. \textbf{(a)} We pre-train a motion tracker on collected imperfect human motion data. \textbf{(b)} We construct a correctable latent action space via online distillation. \textbf{(c)} We train a high-level policy to correct and compose latent actions for tennis task. \textbf{(d)} We transfer the policy to the real world via dynamics randomization and observation noise.}
\label{fig:pipeline}
\vspace{-0.3cm}
\end{figure*}

The overview of LATENT is shown in Figure~\ref{fig:pipeline}. Our method consists of three stages. First, we collect imperfect human motion data (i.e., motion fragments of primitive skills such as forehand stroke, backhand stroke, lateral shuffle, crossover step and so on) with a compact motion capture system in Section~\ref{sec:data_collection}. Then we construct a latent action space that is compatible to potential action corrections with the collected imperfect data in Section~\ref{sec:latent_space}. Finally, we train a high-level policy to correct and compose the primitive skills stored in the latent space to perform tennis skills in Section~\ref{sec:high_level}. 

We use PPO~\cite{ppo} as our reinforcement learning framework and MuJoCo JAX~\cite{todorov2012mujoco,zakka2025mujoco} for simulation in training. We train our policy in parallel on 8 GPUs. Our high-level planner and low-level controller run at 50 Hz, and the simulation frequency is set to 2000 Hz to accurately model ball-racket and ball-ground contact.

\subsection{Collection of Imperfect Human Motion Data} 
\label{sec:data_collection}
Although collecting comprehensive human-tennis motion data from real-world tennis matches is difficult and costly due to the sport's large movement range, multi-rally nature and high precision requirements, acquiring the primitive action skills involved is considerably more feasible. 

We invited five amateur tennis players to collect primitive skills commonly used in tennis, including forehand stroke, backhand stroke, lateral shuffle, crossover step and so on, within an optical motion capture system. Since we collect only primitive skill demonstrations rather than complete motion sequences from tennis matches, the requirements on the motion capture system are significantly reduced. In our setting, the system covers an effective area of only $3 m \times 5 m$, which is more than $17\times$ smaller than a full-size tennis court, resulting in substantial savings in both capture area and the number of cameras compared to collecting complete human motion data in tennis. 

In total, we collect five hours of motion data, without performing any editing or annotation. After collection, we retarget the human motion to humanoid motion with LocoMuJoCo~\cite{al2023locomujoco}.

\subsection{Correctable Latent Action Space Construction}
\label{sec:latent_space}
To learn a latent space of the primitive skills, we first train a motion tracker to imitate the collected motion data and then distill it to a latent model with a variational information bottleneck.

\subsubsection{Motion tracker Pre-training}
\label{sec:tracker_pretrain}
To transform noisy humanoid motion data to executable humanoid primitive skills, we train a motion tracker to imitate the collected motion data. We use the motion tracking framework of Any2Track~\cite{zhang2025track}. 
% Differently, the tracker only outputs motions excluding the wrist (noted as $a^{\text{body}}$, and this design will be discussed in detail in Section \ref{sec:high_level}.
Our tracker can be seen as an RL policy \(\pi_{\text{tracker}}(\hat{a}^{\text{body}}_{t}|s_{t}, \tilde s_{t+1})\), which maps humanoid proprioception state and motion target to low-level robot actions. At each timestep $t$, the inputs of policy $\pi_{\text{tracker}}$ consist of current state $s_{t}$ and current motion target $\tilde s_{t+1}$ from the collected motion data. The current state $s_{t}$ includes angular velocity, projected gravity, per-joint position, per-joint velocity, and last-frame action. The policy $\pi_{\text{tracker}}$ needs to output joint actions $\hat{a}^{\text{body}}_{t}$, which is further fed into a PD controller to compute actuator torques. The task goal of motion tracking is that at timestep ${t+1}$, the robot's state $s_{t+1}$ should be as close as possible to the target motion depicted by $\tilde s_{t+1}$ while maintaining balance and safety.

As mentioned before, our collected motion data do not contain precise wrist motions during striking a tennis ball, and errors are further amplified during human-to-humanoid retargeting. Therefore, in the subsequent steps, the high-level policy needs to provide corrective adjustments to the wrist joints. To enable the controller to robustly handle these potential corrections while preserving stability, we remove the control signals about right wrists (racket-holding) in $\hat{a}^{\text{body}}_{t}$ during training and apply additional random perturbations on the right wrist joints. The tracker needs to imitate the motions of the remaining joints against the unknown wrist disturbances.

\subsubsection{Online Distillation with Variational Bottleneck}
\label{sec:online_distillation}
After learning to imitate the collected human motion fragments, the tracker now contains sufficient primitive skills to play tennis. To learn a compact and structured representation of these primitive skills, we train a student network with a latent representation via online distillation~\cite{luo2023universal, zhang2025unleashing, lu2025habitizing}. For the student network, we adopt an encoder-decoder framework with a conditional variational information bottleneck to encode skills in a continuous skill space, which includes a posterior variational encoder $\mathcal{E}(z^{q}_t | s_t, \tilde s_{t+1}) = \mathcal{N}(z^{q}_t; \mu^e(s_t, \tilde s_{t+1}), \sigma^e(s_t, \tilde s_{t+1}))$ to model the latent code distribution conditioned on current state and motion target, a decoder $\mathcal{D}(\hat{a}^{\text{body}}_t | s_t, z^{q}_t)$ maps the sampled latent code to action conditioned on state.

During training, we encode $(s_{t}, \tilde{s}_{t+1})$ via $\mathcal{E}$ and sample latent code $z^{q}_t \sim \mathcal{N}(z^{q}_t; \mu^e, \sigma^e)$. Then we use the decoder to decode $z^{q}_t$ into humanoid actions $a_{t}^{\text{body}}$. We train the networks to reconstruct the action of the teacher tracker $\hat{a}_{t}^{\text{body}}$ and keep the latent distribution to be close to the prior.
The total loss can be written as:
\begin{equation}
\mathcal{L} = \lambda_{1} \mathcal{L}_{\text{action}} + \lambda_{2} \mathcal{L}_{\text{KL}},
\end{equation}
where
\begin{equation}
\mathcal{L}_{\text{action}} = \mathbb{E}_{(s_t, \hat{a}^{\text{body}}_t) \sim \mathcal{D}_{\text{agg}}} \left[ \| \hat{a}^{\text{body}}_{t} - a_{t}^{\text{body}} \|^2 \right] 
\end{equation}
is the supervision loss of DAgger~\cite{ross2011reduction}, an online distillation method. $\mathcal{D}_{\text{agg}}$ represents the data buffer consist of student current state $s_t$ and teacher action $\hat{a}_t^{\text{body}}$. Meanwhile, $a_t^{\text{body}}$ denotes the action from the student policy. As mentioned before, $\hat{a}^{\text{body}}_{t}$ and $a_{t}^{\text{body}}$ do not contain any control signals for the right wrist. We maintain disturbances on the right wrist joints during online distillation so that the latent space becomes robust to such perturbations.

Rather than training the networks against a fixed unimodal Gaussian as commonly used by VAEs, we utilize a learnable conditional prior $\mathcal{P}(z^{p}_t | s_t) = \mathcal{N}(z^{p}_t; \mu^p(s_t), \sigma^p(s_t))$ to capture state-based action distribution since the action distributions of different robot states differ significantly (e.g., lateral shuffle versus racket-swinging for ball striking). The KL-divergence loss:
\begin{equation}
\mathcal{L}_{\text{KL}} = D_{\text{KL}}(\mathcal{E}(z^{q}_t | s_t, \tilde{s}_{t+1}) \parallel \mathcal{P}(z^{p}_t | s_t))
\end{equation}
encourages the encoded posterior distribution to be close to the learnable prior.

\subsection{High-Level Policy Learning}
\label{sec:high_level}
We train a high-level policy $\pi_{\text{planner}}$ to correct and compose the primitive skills stored in the pre-constructed latent action space to achieve athletic humanoid tennis behavior. At each timestep $t$, the inputs of the high-level policy $\pi_{\text{planner}}$ consist of current robot state $s_{t}$, global information about the robot root pose $g_{t}$, and ball state $b_{t}$. 

\subsubsection{Task Setting} 
\begin{figure}[h]
    \centering
    % \vspace{0.2cm}
    \includegraphics[width=1.0\linewidth]{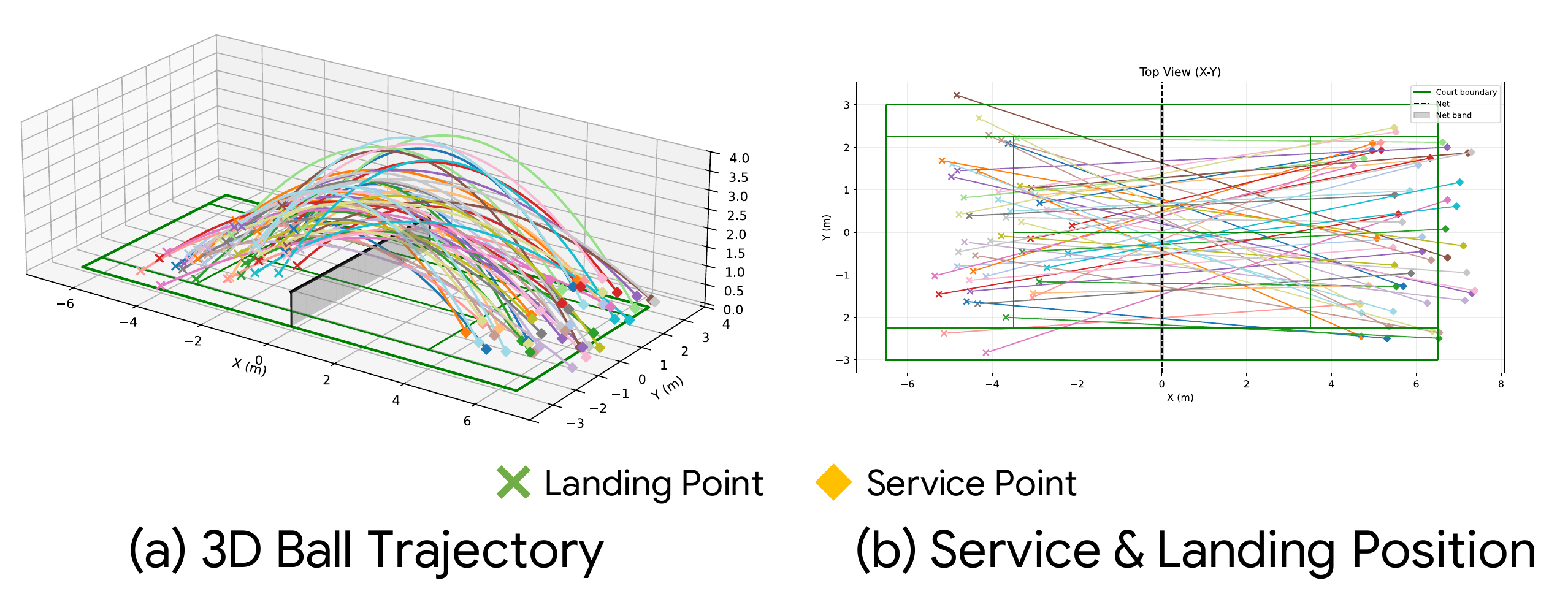}
    \caption{\textbf{Visualization of generated ball trajectories.}}
    \label{fig:trajectory}
% \vspace{-0.2cm}
\end{figure}

At the beginning of each episode, the humanoid robot is randomly initialized at a position on the court in a ready pose. In each episode, the robot is required to consecutively return eight tennis balls to the target location, each initialized with different positions and velocities, to the target location. The generated ball trajectories in our work is visualized in Figure~\ref{fig:trajectory}. The incoming balls are launched every 2 seconds. The rewards are shown in Table~\ref{table:rewards}.

\begin{table}[]
\vspace{0.2cm}
\centering
\footnotesize
\caption{\textbf{Reward terms used in LATENT.} The reward terms are divided into three types: Task, Regularization, and Termination.}
\resizebox{1.0\linewidth}{!}{
\setlength{\tabcolsep}{2pt}
\begin{tabular}{@{}cccc@{}}
\toprule
Term & Weight & Term & Weight \\ \midrule

\multicolumn{4}{c}{Task} \\ \midrule

Approach to ball & $10.0$ & Ball landing & $30.0$ \\

Hit success & $200.0$ &  &  \\

\midrule
\multicolumn{4}{c}{Regularization} \\ \midrule

High-level action & $-5 \cdot 10^{-3}$ & Torque penalty & $-2 \cdot 10^{-5}$ \\

Lower body action rate & $-1.0$ & Whole body action rate & $-0.5$ \\

Racket acceleration & $-1 \cdot 10^{-6}$ & Joint smoothness & $-2 \cdot 10^{-6}$ \\

Correction action & $-5.0$ & Correction action rate & $-1.0$ \\

Wrist torque & $-4 \cdot 10^{-5}$ & Wrist joint smoothness & $-4 \cdot 10^{-6}$ \\

Joint position limit & $-10.0$ & Joint velocity limit & $-5.0$ \\

Self-collision & $-10.0$ & Net clearance & $100.0$ \\

Ball velocity constraint & $50.0$ & Racket velocity constraint & $50.0$ \\

Pelvis facing forward & $1.0$ & & \\

\midrule
\multicolumn{4}{c}{Termination} \\ \midrule

Fall & $-200.0$ & Miss ball & $-200.0$ \\

Ball-net collision & $-50.0$ &  Ball out of bounds & $-50.0$ \\

Stroke style violation & $-50.0$ &  & \\

\bottomrule
\end{tabular}}
\label{table:rewards}
\vspace{-0.3cm}
\end{table}

\subsubsection{Action Correction}
Accurately capturing the subtle wrist motions of human athletes during racket swings is extremely challenging. The embodiment gap between humans and humanoids further amplifies this discrepancy. As a result, the retargeted wrist motions are highly inaccurate and cannot be directly utilized. To mitigate this issue, we adopt a hybrid control approach~\cite{zhang2023learning}. The high-level policy simultaneously outputs $a^{\text{planner}}_t=[a^{\text{latent}}_{t}, a^{\text{correct}}_{t}]$, including latent actions and direct control commands for the right wrist. Since the latent action space is designed to be robust to right wrist corrections, the high-level policy can output arbitrary wrist motion adjustments without compromising the stability and agility of the lower-body movements.

\subsubsection{Latent Action Barrier}
The learned latent space creates a compact and continuous manifold of the primitive skills characterized by the pre-collected human motion data. We aim to learn a high-level policy that can correct and compose these primitive skills to accomplish the tennis return task, thereby enabling the policy to complete the task while preserving the motion naturalness. However, since our imperfect human motion data only provides priors about natural motor skills and does not contain any knowledge about how to utilize them for the tennis return task, the high-level policy needs to discover how to leverage these skills for task accomplishment via reinforcement learning. A key challenge lies the balance between task performance and adherence to the primitive skill priors during high-level policy training. Although the latent space contains many high-quality primitive skills, without proper constraints, the high-level policy can easily exploit this space during RL training, sampling suboptimal actions and composing them into unnatural, jittery motion sequences. For example, the high-level policy may switch back and forth between different locomotion primitives while running toward the incoming ball, resulting in low-quality motions that nevertheless succeed in completing the task.

\begin{figure}[h]
    \centering
    % \vspace{0.2cm}
    \includegraphics[width=0.95\linewidth]{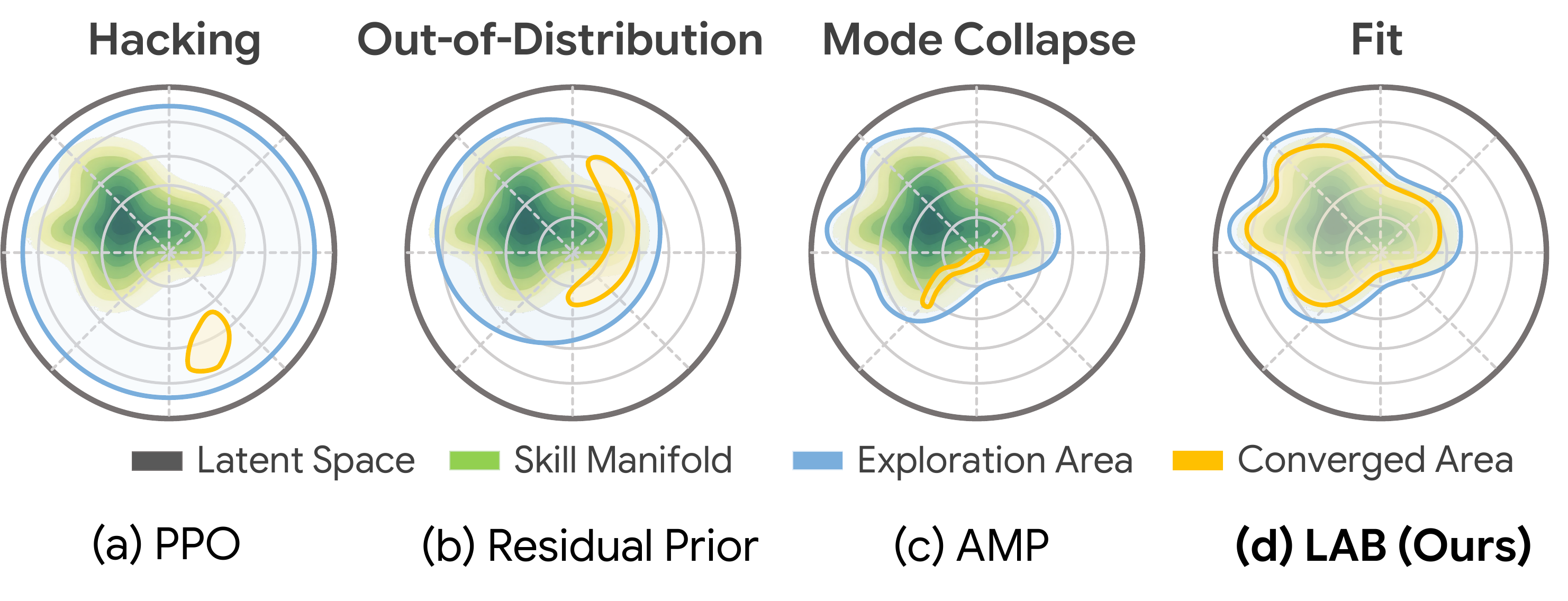}
    \vspace{0.05cm}
    \caption{\textbf{The motivation of Latent Action Barrier (LAB).}}
    \label{fig:motivation}
    \vspace{-0.1cm}
\end{figure}

To address this problem, we propose to constrain the exploration space of the high-level policy. The learnable conditional prior distribution $\mathcal{P}$ depicts state-dependent action distribution and can serve as a good starting point. The action space of $a^{\text{latent}}_{t}$ can be formed as a residual action space~\cite{luo2023universal, zhang2025unleashing} with respect to the prior's mean $\mu^p_{t}$.

However, the manifold of primitive skills is not solely determined by the prior mean, but is also shaped by the associated variance. If the exploration space of the residual latent action $a^{\text{latent}}_{t}$ is not constrained, the high-level policy can still easily sample latent actions that deviate significantly from the prior's mean $\mu^p_{t}$, ultimately decoding them into suboptimal motor actions. Therefore, we propose to constrain the exploration range of the residual latent action $a^{\text{latent}}_{t}$ within a Latent Action Barrier around the prior's mean $\mu^p_{t}$. In our setting, we observe that the prior's standard deviation $\sigma^{p}_{t}$ varies significantly across different states, as well as across different dimensions of the latent space. Thus, we design the barrier based on the Mahalanobis distance rather than the Euclidean distance (i.e., the barrier range is adaptively scaled according to the state-dependent standard deviation). The final PD target action can be written as:
\begin{equation}
a^{\text{full}}_{t}=[\mathcal{D}(s_{t},\mu^{p}_{t} + \lambda \sigma^{p}_{t} \cdot tanh(a^{\text{latent}}_{t})), a^{\text{correct}}_{t}],
\end{equation}
where the hyperparameter $\lambda$ is a scale factor to control the exploration range. Figure~\ref{fig:motivation} demonstrates the intuitive motivation behind Latent Action Barrier (LAB).

\section{Sim-to-Real Transfer}
Due to the high dynamics of humanoid movement, the high speed of the tennis ball, and the high precision of humanoid-tennis interaction, transferring the learned policy from simulation to the real world is non-trivial. We will first introduce our dynamics randomization techniques in Section~\ref{sec:dynamics_randomization}. Then we will introduce the observation noise in Section~\ref{sec:data_obs_noise}. 
At last, we will introduce our real-world deployment in Section~\ref{sec:deployment}.

\subsection{Dynamics Randomization}
\label{sec:dynamics_randomization}
To mitigate the dynamics gap between simulation and the real world, we introduce dynamics randomization of the humanoid and the tennis ball, as shown in Table~\ref{table:dynamics}.

For the robot, we introduce randomization in friction, armature, mass, and center of mass.

For the tennis ball, a major challenge lies in modeling contacts between the ball and the ground or racket, as well as the aerodynamic effects on ball. These factors can significantly influence ball trajectories and constitute the primary sources of the sim-to-real gap. Existing works~\cite{su2025hitter, zhang2023learning} often rely on precise system identification and explicit modeling of complex aerodynamic effects. However, we found that sufficient domain randomization in dynamics enables effective sim-to-real transfer and greatly improve the robustness of the system (Table \ref{table:dynamics}). We introduce randomization in ball mass, ball-ground restitution, tangential damping, and the air-drag coefficient $k$. The air-drag force is modeled as $\boldsymbol{f}_{air} = - k * m * \boldsymbol{v} * ||\boldsymbol{v}||$, where $m$ is the mass of the object and $\boldsymbol{v}$ denotes its velocity.

\begin{table}[]
\centering
\caption{\textbf{Dynamics randomization used in LATENT.} We randomize robot dynamics and tennis ball physics.}
\vspace{0.2cm}
\resizebox{0.75\linewidth}{!}{
\begin{tabular}{@{}ll@{}}
\toprule
Term & Value \\ 
\midrule
\multicolumn{2}{c}{Robot} \\ 
\midrule
Feet friction                 & $\mathcal{U}(1.0, 2.5)$ \\
Joint friction        & $\times \mathcal{U}(0.75, 1.25)$ \\
Armature              & $\times \mathcal{U}(1.0, 1.05)$ \\
Body mass       & $\times \mathcal{U}(0.75, 1.25)$ \\
Body CoM  & $\mathcal{U}(-0.05, 0.05)$ \\
Racket CoM  & $\mathcal{U}(-0.1, 0.1)$ \\
Racket mass           & $\times \mathcal{U}(0.8, 1.2)$ \\

\midrule
\multicolumn{2}{c}{Tennis Ball} \\ 
\midrule
Ball mass              & $\times \mathcal{U}(0.95, 1.05)$ \\
Ball-ground restitution & $\mathcal{U}(0.71, 0.79)$ \\
Ball-ground tangential damping & $\mathcal{U}(0.4, 0.7)$ \\
Ball air-drag coefficient    & $\mathcal{U}(0.01, 0.04)$ \\

\bottomrule
\end{tabular}
}
\label{table:dynamics}
\vspace{-0.3cm}
\end{table}

\subsection{Observation Noise}
\label{sec:data_obs_noise}

To mitigate observation errors caused by inaccurate real-world state estimation and system latency, we introduce our modeling of the observation noise in the simulation.

We add uniform noise, frame-level dropouts, and latency in the policy observations. Due to the high speed nature of tennis, accurately estimating the ball’s state is crucial for successful ball striking. We first obtain the ball position and then compute its velocity via finite differencing. However, we observe that even small amounts of noise in position estimation can significantly degrade the accuracy of the inferred velocity. To mitigate this issue, we maintain a four-frame sliding window in both simulation and the real world, and use the averaged velocity within this window as the ball observation.

\subsection{Real-world Deployment}
\label{sec:deployment}
In the real world, we attach a tennis racket to the humanoid’s right wrist using a 3D-printed connector and replace the right hand. To obtain the global state of the robot and the tennis ball, we utilize an optical motion capture system. We attach multiple reflective markers to the robot base and model them as a rigid body to estimate its global 6D pose. The ball is covered with reflective tape and modeled as a passive reflective marker.

\section{Experiments}
\begin{table*}
\centering
\renewcommand{\arraystretch}{1.2}
\caption{\textbf{Comparison with baseline methods.} \textbf{Bold} numbers indicate the best performance.}
\label{tab:main_comparison}
\setlength{\tabcolsep}{4pt}
\resizebox{\textwidth}{!}{
\begin{tabular}{lccccccccccccccccccc}
\hline
\multirow{2}{*}{Method} & \multicolumn{4}{c}{\textbf{\color[rgb]{0,0.353,0.588}{Forehand}}}   & \multicolumn{1}{l}{\multirow{2}{*}{}} & \multicolumn{4}{c}{\textbf{\color[rgb]{0,0.353,0.588}{Backhand}}}    & \multicolumn{1}{l}{\multirow{2}{*}{}} & \multicolumn{4}{c}{\textbf{\color[rgb]{0.784,0.431,0}{Forecourt}}}     & \multicolumn{1}{l}{\multirow{2}{*}{}} & \multicolumn{4}{c}{\textbf{\color[rgb]{0.784,0.431,0}{Backcourt}}} \\ \cline{2-5} \cline{7-10} \cline{12-15} \cline{17-20} 
                        & SR↑           & DE↓         & Smth↓   & Torque↓     & \multicolumn{1}{l}{}                  & SR↑           & DE↓         & Smth↓   & Torque↓        & \multicolumn{1}{l}{}                  & SR↑           & DE↓         & Smth↓   & Torque↓         & \multicolumn{1}{l}{}                  & SR↑           & DE↓         & Smth↓   & Torque↓          \\ \hline
PPO~\cite{ppo}                     & N/A          & N/A           & N/A      &  N/A    &                                       & N/A          & N/A         & N/A     &   N/A    &                                       & N/A           & N/A           & N/A  &   N/A      &                                       & N/A             & N/A             & N/A     &    N/A     \\
MotionVAE~\cite{zhang2023learning}                     & N/A          & N/A          & N/A     & N/A     &                                        & N/A          & N/A          & N/A     & N/A     &                                        & N/A          & N/A          & N/A     & N/A     &                                         & N/A          & N/A          & N/A     & N/A         \\
AMP~\cite{peng2021amp}           & 41.32          & 4.71          & 46.52    &    13.79  &                                       & 36.81          & 5.22          & 37.68   &  15.74     &                                       & 49.48          & 4.20          & 36.16  &    18.26     &                                       & 41.73             & 4.85            & 41.77    &   19.20     \\
ASE~\cite{peng2022ase} &
63.47 & 3.62 & 38.91 & 11.84 &
&
55.22 & 4.01 & 31.75 & 12.92 &
&
71.36 & 3.28 & 30.84 & 14.05 &
&
68.59 & 3.90 & 35.12 & 15.27 \\
PULSE~\cite{luo2023universal} &
71.85 & 3.01 & 32.44 & 9.76 &
&
63.38 & 3.15 & 29.60 & 10.83 &
&
74.52 & 2.97 & 28.11 & 12.01 &
&
72.94 & 3.02 & 31.27 & 13.44 \\
\textbf{Ours}        & \textbf{96.52} & \textbf{1.32} & \textbf{25.61} & \textbf{7.40} &    & \textbf{82.10} & \textbf{1.89} & \textbf{27.76} &  \textbf{7.71} & & \textbf{86.35} & \textbf{1.73}          & \textbf{25.51} & \textbf{7.47} & & \textbf{89.80}    & \textbf{1.55}   & \textbf{27.20} & \textbf{7.60} \\ \hline
\end{tabular}}
\vspace{-0.2cm}
\end{table*}

\begin{table*}
\centering
\renewcommand{\arraystretch}{1.2}
\caption{\textbf{Ablation study.} \textbf{Bold} numbers indicate the best performance.}
\label{tab:ablation}
\setlength{\tabcolsep}{4pt}
\resizebox{\textwidth}{!}{
\begin{tabular}{lccccccccccccccccccc}
\hline
\multirow{2}{*}{Method} & \multicolumn{4}{c}{\textbf{\color[rgb]{0,0.353,0.588}{Forehand}}}   & \multicolumn{1}{l}{\multirow{2}{*}{}} & \multicolumn{4}{c}{\textbf{\color[rgb]{0,0.353,0.588}{Backhand}}}    & \multicolumn{1}{l}{\multirow{2}{*}{}} & \multicolumn{4}{c}{\textbf{\color[rgb]{0.784,0.431,0}{Forecourt}}}     & \multicolumn{1}{l}{\multirow{2}{*}{}} & \multicolumn{4}{c}{\textbf{\color[rgb]{0.784,0.431,0}{Backcourt}}} \\ \cline{2-5} \cline{7-10} \cline{12-15} \cline{17-20} 
                        & SR↑           & DE↓         & Smth↓   & Torque↓     & \multicolumn{1}{l}{}                  & SR↑           & DE↓         & Smth↓   & Torque↓        & \multicolumn{1}{l}{}                  & SR↑           & DE↓         & Smth↓   & Torque↓         & \multicolumn{1}{l}{}                  & SR↑           & DE↓         & Smth↓   & Torque↓          \\ \hline
w/o Corr. &
82.36 & 3.41 & 26.72 & 8.85 &
&
68.94 & 3.03 & 29.55 & 8.92 &
&
74.21 & 3.18 & 26.86 & 9.64 &
&
79.05 & 3.52 & 29.94 & 9.38 \\
w/o LAB &
93.12 & 2.17 & 37.64 & 12.53 &
&
76.96 & 2.21 & 31.77 & 12.11 &
&
84.73 & 1.96 & 30.59 & 10.94 &
&
87.44 & 2.05 & 32.08 & 13.62 \\
\textbf{Ours}        & \textbf{96.52} & \textbf{1.32} & \textbf{25.61} & \textbf{7.40} &    & \textbf{82.10} & \textbf{1.89} & \textbf{27.76} &  \textbf{7.71} & & \textbf{86.35} & \textbf{1.73}          & \textbf{25.51} & \textbf{7.47} & & \textbf{89.80}    & \textbf{1.55}   & \textbf{27.20} & \textbf{7.60} \\ \hline
\end{tabular}}
\vspace{-0.2cm}
\end{table*}
In this section, we provide extensive experimental results in both the MuJoCo~\cite{todorov2012mujoco} simulator and the real-world deployment. We choose the 29-DoF Unitree G1 humanoid robot for all our experiments. Here, we aim at addressing the following three questions:

\begin{itemize}
    \item \textbf{Q1}: How does LATENT perform on the tennis return task compared with existing methods?
    \item \textbf{Q2}: How does each component contribute to the final performance of LATENT?
    \item \textbf{Q3}: How does LATENT perform in the real world?
\end{itemize}

\subsection{Comparison of Performances}
\label{sec:exp_main}
To address \textbf{Q1} (\textit{How does LATENT perform on the tennis return task compared with existing methods?}), we compare LATENT with baseline methods on a simulated tennis return task.

\subsubsection{Experiment Setting}
In this experiment, we evaluate the humanoid’s performance of the tennis return task. We use the method described in Section~\ref{sec:high_level} to randomly initialize the ball and the robot. We evaluate the performance of our method under different stroke types (forehand and backhand) and across different court regions (forecourt and backcourt). The results are averaged over a total of 10{,}000 trials.
\subsubsection{Experiment Metrics}
We use the following two metrics to measure the task performance:
\begin{itemize}
    \item \textbf{Success Rate (SR)} records the percentage of trials that humanoids successfully return an incoming tennis ball to within 2.5 meters of the target location. 
    \item \textbf{Distance Error (DE)} is the averaged distance between the returned tennis ball’s actual landing position and the target location.
\end{itemize}
We also use the following two metrics to measure the action quality of the humanoid:

\begin{itemize}
    \item \textbf{Smoothness (Smth)} is the averaged acceleration of every joint to measure the naturalness and stability of the humanoid actions.
    \item \textbf{Torque} is the averaged torque of every joint to measure the physical feasibility and hardware safety of the humanoid actions.
\end{itemize}
\subsubsection{Baselines}
We choose the following methods as baselines:
\begin{itemize}
    \item \textbf{PPO}~\cite{ppo}: We directly train the policy with the vanilla PPO algorithm to accomplish the tennis return task without any human motion prior.
    \item \textbf{MotionVAE}~\cite{zhang2023learning}: We adopt the MotionVAE framework used in Vid2Player3D, where we train a kinematics-based motion VAE. The high-level policy is trained to sample in the latent motion space. The sampled latent is decoded into kinematic motions, which are then tracked by a decoupled motion tracker.
    \item \textbf{AMP}~\cite{peng2021amp}: We adopt AMP, a reinforcement learning framework that incorporates human motion prior through adversarial training.
    \item \textbf{ASE}~\cite{peng2022ase}: We adopt ASE, a method that learns reusable adversarially trained skill embeddings to accomplish high-level tasks. 
    \item \textbf{PULSE}~\cite{luo2023universal}: We adopt PULSE, which learns VAE-style motion representation through online distillation from a pre-trained motion tracker.  The key difference from our method is that, in this baseline, we strictly follow the original paper’s procedure to construct and sample from the latent space and do not apply any wrist corrections.
\end{itemize}
\subsubsection{Experiment Results}
The results are shown in Table~\ref{tab:main_comparison}. ``N/A'' indicates that the method fails to converge. LATENT outperforms baseline methods in tennis return task performance and action quality. For vanilla PPO, it is extremely difficult to learn tennis skills from scratch without any motion prior. We implement the MotionVAE proposed in Vid2Player3D. However, we found that the decoupled design of the motion generator and the tracker places stringent requirements on both components. Low-quality motions generated by the motion generator can make accurate tracking difficult, which in turn further degrades the quality of the autoregressively generated kinematic motions. Such accumulated errors can quickly cause the robot to lose balance and fall. Vid2Player3D~\cite{zhang2023learning} alleviates this issue by introducing residual force and relying on unrealistic physics assumptions. However, the requirement of real-robot deployment prevents us from adopting such assumptions.
For AMP and ASE, we found that these adversarial learning-based methods struggle to strike a balance between task completion and adherence to the motion prior. For PULSE,  we found that its approach to constructing and utilizing the latent action space leads to lower-quality sampled actions. Moreover, the absence of wrist-level correction further degrades its task completion performance.

% \begin{figure}
%     \centering
%     \vspace{0.1cm}
%     \includegraphics[width=1.0\linewidth]{figures/realworld.pdf}
%     \caption{\textbf{Analysis of 1M shots.}}
%     \label{fig:realworld}
% \vspace{-0.5cm}
% \end{figure}

\subsection{Ablation of Key Components}

To address \textbf{Q2} (\textit{How does each component contribute to the final performance of LATENT?}), we conduct ablation studies on each key component of our method.

\subsubsection{Experiment Setting and Metrics}
We adopt the same experiment setting and metrics in Section~\ref{sec:exp_main}.
\subsubsection{Baselines}
We choose the following methods as baselines:
\begin{itemize}
    \item \textbf{Ours w/o Correction (w/o Corr.)}: We train a whole-body motion tracker, including the right wrist joints, and then distill it to a latent space for all joints. The high-level policy is not allowed to correct the right wrist actions.
    \item \textbf{Ours w/o Latent Action Barrier (w/o LAB)}: We remove the latent action barrier and allow the high-level policy to output residual latent actions in an unconstrained manner.
\end{itemize}
\subsubsection{Experiment Results}
The results are shown in Table~\ref{tab:ablation}. Our key designs contribute to the final performance. Without wrist correction, the policy's success rate drops significantly. Without the latent action barrier, the high-level policy tends to exploit the latent space to sample suboptimal actions, resulting in jittery motion sequences and ultimately degrading the overall policy performance.

\subsection{Evaluation of real-world performance}

To address \textbf{Q3} (\textit{How does LATENT perform in the real world?}), we conduct real-world experiments.

\subsubsection{Experiment Setting} We conduct 20 consecutive human-robot rally matches to evaluate our policy’s tennis return performance in the real world, with randomized initial positions and velocities of the ball. For each successfully returned ball, we record the landing position, categorize it into forecourt or backcourt regions, and analyze the usage of forehand and backhand strokes.
\subsubsection{Experiment Metrics} We still adopt \textbf{Success Rate (SR)} and \textbf{Distance Error (DE)} as our metrics. However, due to the sim-to-real gap, we observe that the landing accuracy on the real-world performance is relatively low. Therefore, we consider a trial successful if the robot successfully returns the ball within the opponent's court boundaries.
\subsubsection{Baselines}
Since sim-to-real transfer for the humanoid robot itself has been extensively studied in prior works~\cite{zhang2025track,he2024learning,he2024omnih2o} and is not the primary focus of this paper, we design the baseline mainly to validate the effectiveness of incorporating dynamics randomization and observation noise for the ball.
\begin{itemize}
    \item \textbf{Ours w/o Dynamics Randomization (w/o DR)}: In this baseline, we remove all dynamics randomization applied to the ball.
    \item \textbf{Ours w/o Observation Noise (w/o ON)}: In this baseline, we remove all observation noise applied to the ball.
\end{itemize}
\subsubsection{Experiment Results}

\begin{table}
\centering
\renewcommand{\arraystretch}{1.2}
\caption{\textbf{Real-world performance.} 
\vspace{0.2cm}
\textbf{Bold} numbers indicate the best, and \underline{underlined} numbers indicate the second.}
\label{tab:realworld}
\setlength{\tabcolsep}{3pt}
\resizebox{\columnwidth}{!}{
\begin{tabular}{lccccccccccc}
\hline

\multirow{2}{*}{Method} 
& \multicolumn{2}{c}{\textbf{\color[rgb]{0,0.353,0.588}{Forehand}}}
& \multicolumn{1}{l}{\multirow{2}{*}{}}
& \multicolumn{2}{c}{\textbf{\color[rgb]{0,0.353,0.588}{Backhand}}}
& \multicolumn{1}{l}{\multirow{2}{*}{}}
& \multicolumn{2}{c}{\textbf{\color[rgb]{0.784,0.431,0}{Forecourt}}}
& \multicolumn{1}{l}{\multirow{2}{*}{}}
& \multicolumn{2}{c}{\textbf{\color[rgb]{0.784,0.431,0}{Backcourt}}} \\

\cline{2-3} \cline{5-6} \cline{8-9} \cline{11-12}

& SR↑ & DE↓ &
& SR↑ & DE↓ &
& SR↑ & DE↓ &
& SR↑ & DE↓ \\

\hline

w/o DR
& 16.67\% & 5.60 &
& \underline{25.00\%} & 4.57 &
& 28.57\% & \underline{3.59} &
& 14.29\% & 6.23 \\

w/o ON
& \underline{50.00\%} & \underline{3.28} &
& 0.00\% & N/A &
& \underline{28.57\%} & 3.69 &
& \underline{37.50\%} & \underline{3.97} \\

\textbf{Ours}
& \textbf{90.90\%} & \textbf{3.15} &
& \textbf{77.78\%} & \textbf{3.89} &
& \textbf{88.89\%} & \textbf{3.57} &
& \textbf{81.82\%} & \textbf{3.80} \\

\hline
\end{tabular}}
\vspace{-0.5cm}
\end{table}

The real-world evaluation results are shown in Table~\ref{tab:realworld}. Our method demonstrates strong real-world performance. We show that adequately randomizing ball dynamics and observation can enable effective sim-to-real transfer without relying on precise system identification.
\section{Conclusion}
We propose LATENT, a novel system to learn athletic humanoid tennis skills from imperfect human motion data. We successfully deploy our method on a real-world humanoid robot and can stably sustain multi-shot rallies with human players. Although this work primarily focuses on the tennis return task, the proposed framework has the potential to generalize to a broader range of tasks where complete and high-quality human motion data are unavailable (e.g., soccer and parkour).

The current work presents several potential directions for improvement. First, the proposed method relies on a motion capture system for real-world deployment. Incorporating active vision could help alleviate this limitation. Second, the current task formulation focuses on returning randomly initialized incoming balls to target locations, which still differs from a real two-player tennis match. Achieving a level of performance comparable to professional human players may require introducing a multi-agent training framework.

\bibliographystyle{assets/plainnat}
\bibliography{main}

\appendix
\newpage
\clearpage
\onecolumn

\section{Additional Experiments}
In this section, we provide further analysis of our policy.

\subsection{Robot Movement Coverage}

\begin{figure}[h]
\vspace{-0.2cm}
\centering
\includegraphics[width=0.98\linewidth]{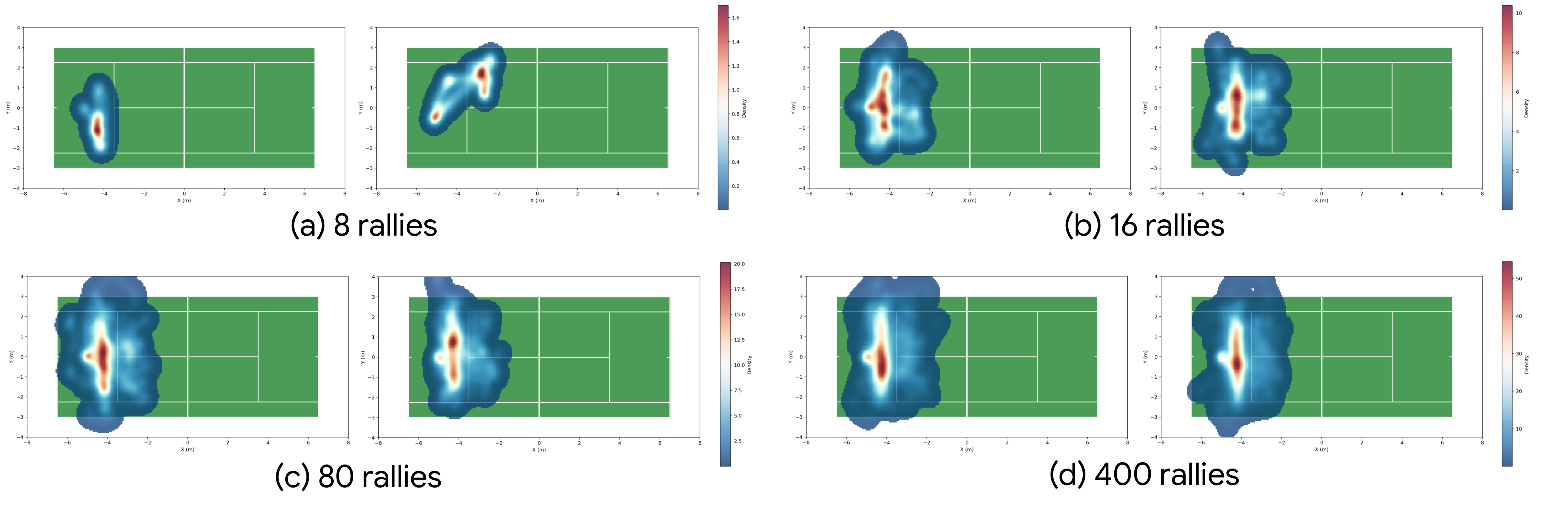}
\caption{\textbf{Robot movement coverage during consecutive ball returns.}
Heatmaps of the robot's global positions accumulated over different numbers of consecutive ball returns (8, 16, 80, and 400). The learned policy enables effective court coverage and adaptive repositioning during consecutive rallies.}
\label{fig:rally_coverage}
\end{figure}

To further analyze the robot's movement behavior during rallies, we study the spatial distribution of the robot's pelvis positions across consecutive ball returns, as is demonstrated in Figure~\ref{fig:rally_coverage}. 
We aggregate the pelvis trajectories from different numbers of consecutive ball returns (8, 16, 80, and 400) to examine how the spatial coverage evolves as more interactions are accumulated. Positions with density lower than 0.01\% are filtered out for better visualization.

From the resulting distributions, we observe that the robot consistently explores a wide region of the court while maintaining a stable movement structure across episodes. Figure~\ref{fig:rally_coverage} (a)-(d) indicate that the robot is capable of dynamically repositioning itself and maintaining broad court coverage during repeated rallies.

\subsection{Robot-Robot Self-Play}
To further evaluate the learned policy, we conduct robot-robot self-play experiments in simulation. In this setting, two instances of the policy are deployed on opposite sides of the court to exchange tennis returns. The results show that the two robots can maintain stable rallies and is able achieve up to 25 consecutive rallies, demonstrating the robustness of the policies trained with LATENT.

\begin{figure}[h]
\centering
\includegraphics[width=0.92\linewidth]{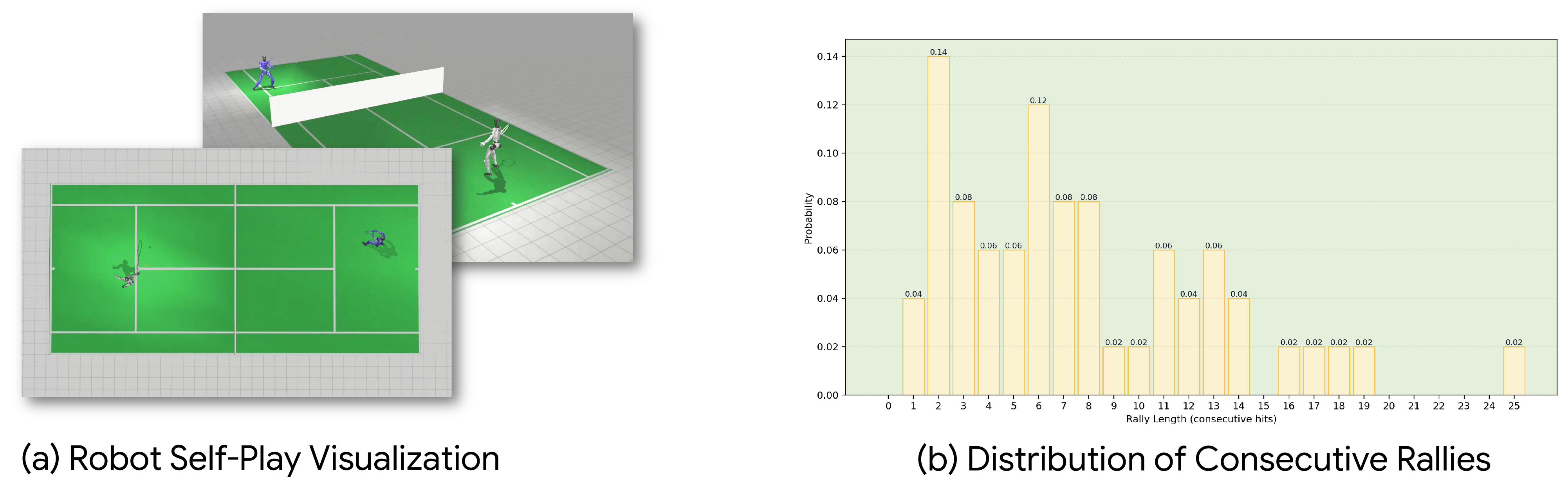}
\caption{\textbf{Robot-robot self-play evaluation in simulation.} 
(a) Visualization of two robots performing self-play on opposite sides of the court. 
(b) Distribution of number of consecutive rallies over 50 random games.}
\label{fig:selfplay}
\vspace{-0.2cm}
\end{figure}

To further improve rally performance between the robot and human players, as well as between robots, future works may incorporate reinforcement learning with self-play. Methods such as Neural Fictitious Self-Play (NFSP)~\cite{heinrich2016deep} and other multi-agent reinforcement learning approaches~\cite{bansal2017emergent, lowe2017multi, foerster2018counterfactual} provide a potential framework for iterative policy improvement through competitive interaction.

\section{Additional Visualizations}
In this section, we provide additional qualitative visualization for both simulation and real-world deployment. 

\subsection{Training in Simulation}

\begin{figure}[h]
\centering
\includegraphics[width=0.95\linewidth]{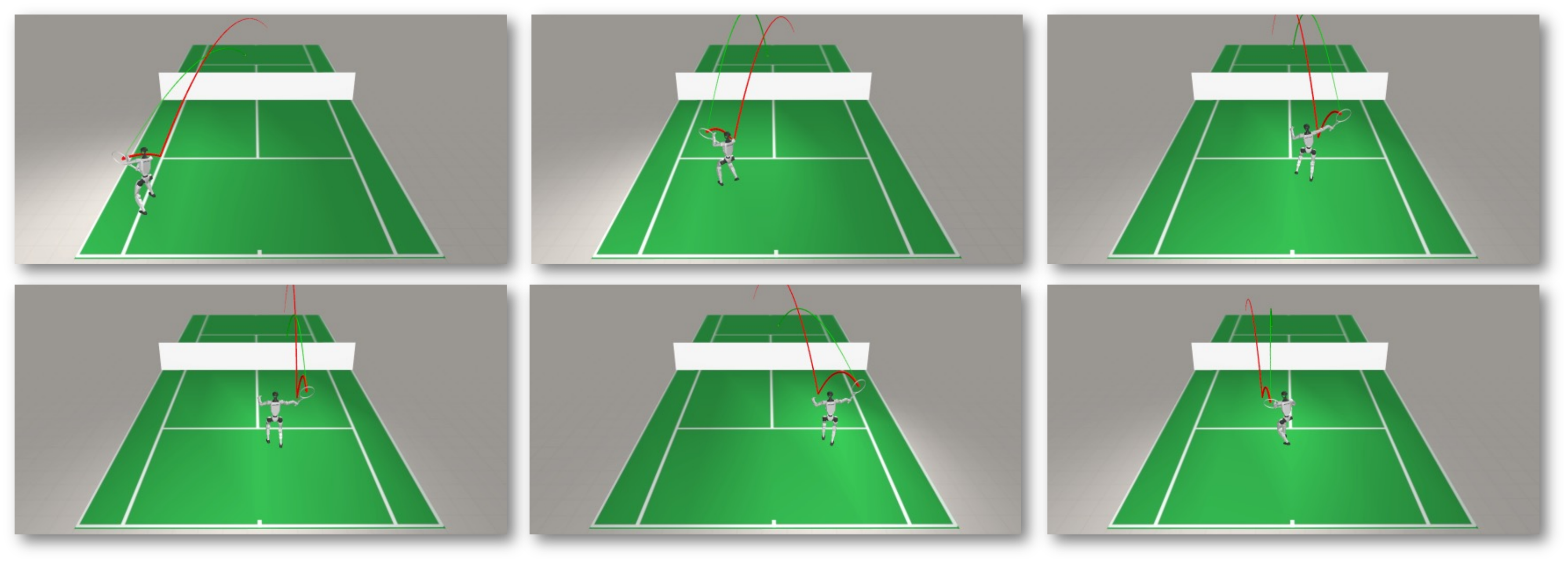}
\caption{\textbf{Different hitting events in simulation.}
The red curves denote incoming ball trajectories, the robot posture corresponds to the moment of ball contact, and the green curves indicate the trajectories of the returned balls.}
\label{fig:simulation}
\end{figure}

Figure~\ref{fig:simulation} shows representative hitting events in simulation. The red curves denote incoming ball trajectories (e.g., serves or opponent shots). The robot posture corresponds to the moment of ball contact. The green curves indicate the resulting return trajectories after the hit. These examples illustrate that the learned policy can adapt its body posture and racket motion to intercept diverse incoming trajectories and generate consistent return shots.

\subsection{Close-up Views of the Robot During Rallies in the Real World}

In our experiments, we build our system based on Unitree G1 and introduce several hardware modifications to improve robustness during motion capture and interaction. Light-diffuse surfaces are added to the robot body to improve motion capture system stability and eliminate spurious reflective areas. A standard tennis racket is mounted to the robot’s right end-effector using a 3D-printed adapter (Figure~\ref{fig:pipeline}). In addition, the end-effector joint connectors are redesigned to improve structural stability during dynamic motions.

\begin{figure}[h]
\centering
\includegraphics[width=0.95\linewidth]{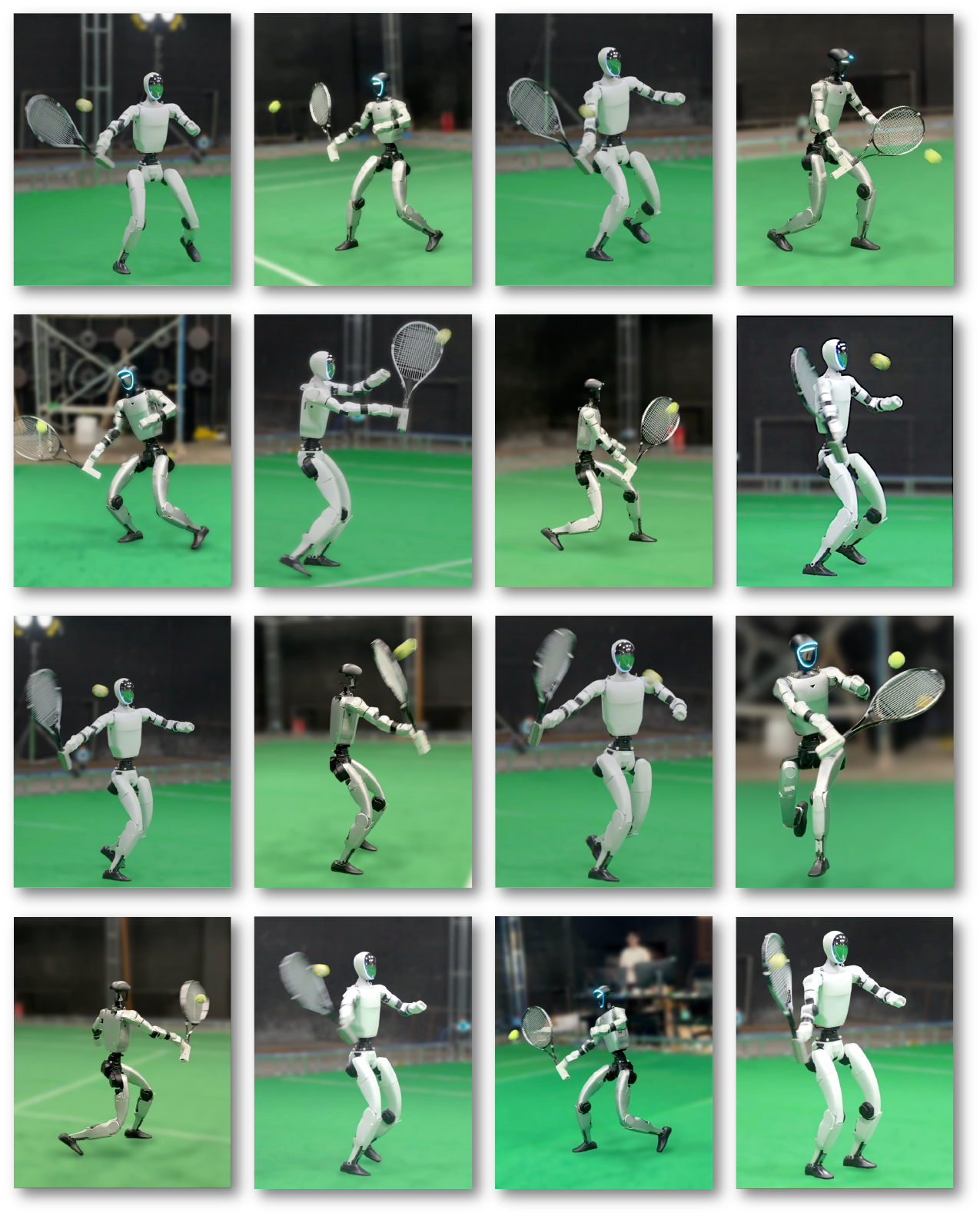}
\caption{\textbf{Close-up real-world examples.}
Representative frames from real-world rallies showing diverse tennis return behaviors, including forehand and backhand strokes, dynamic footwork, and coordinated whole-body motion.}
\label{fig:close_up}
\end{figure}

As shown in Figure~\ref{fig:close_up}, the robot exhibits diverse motion patterns when interacting with incoming balls. Across different rallies and ball trajectories, it adapts its body posture and swing timing to reach the ball and execute stable returns. These behaviors resemble those of human tennis players, where the body dynamically adjusts to maintain balance and achieve effective ball contact. Generally, the learned policy produces robust and natural motion strategies in the real world, enabling the robot to perform dynamic tennis returns under varying conditions.

\end{document}